\pdfoutput=1

\documentclass[11pt]{article}

\usepackage{acl}
\usepackage{amssymb}

\usepackage{times}
\usepackage{latexsym}
\usepackage{tikz}
\usetikzlibrary{positioning,shapes,arrows,calc}
\usetikzlibrary{shapes.geometric, positioning, arrows, calc}
\tikzstyle{process} = [rectangle, rounded corners, minimum width=2.5cm, minimum height=1cm, text centered, draw=black, fill=blue!30]
\tikzstyle{data} = [rectangle, minimum width=2.5cm, minimum height=1cm, text centered, draw=black, fill=green!30]
\tikzstyle{decision} = [diamond, minimum width=2.5cm, minimum height=1cm, text centered, draw=black, fill=red!30]
\tikzstyle{arrow} = [thick,->,>=stealth]
\usepackage{pgfplots}
\pgfplotsset{compat=1.18}
\usetikzlibrary{positioning,shapes,arrows,calc}
\usetikzlibrary{shapes.geometric, positioning}
\usepackage[T1]{fontenc}

\usepackage[utf8]{inputenc}

\usepackage{microtype}
\usepackage{amsmath}
\usepackage{inconsolata}

\usepackage{graphicx}

\title{When retrieval outperforms generation: Dense evidence retrieval for scalable fake news detection\thanks{Published in \textit{Proceedings of the 5th Conference on Language, Data and Knowledge (LDK 2025)}, pages 255--265. \url{https://aclanthology.org/2025.ldk-1.26/}}}

\author{
  Alamgir Munir Qazi$^{1}$ \quad
  John P. McCrae$^{2}$ \quad
  Jamal Abdul Nasir$^{1}$ \\
  $^{1}$School of Computer Science, University of Galway, Ireland \\
  $^{2}$Research Ireland Insight Centre and ADAPT Centre, University of Galway, Ireland \\
  \texttt{\{a.qazi1,jamal.nasir\}@universityofgalway.ie, john@mccr.ae}
}

\begin{document}
\maketitle
\begin{abstract}

The proliferation of misinformation necessitates robust yet computationally efficient fact verification systems. While current state-of-the-art approaches leverage Large Language Models (LLMs) for generating explanatory rationales, these methods face significant computational barriers and hallucination risks in real-world deployments. We present DeReC (Dense Retrieval Classification), a lightweight framework that demonstrates how general-purpose text embeddings can effectively replace autoregressive LLM-based approaches in fact verification tasks. By combining dense retrieval with specialized classification, our system achieves better accuracy while being significantly more efficient. DeReC outperforms explanation-generating LLMs in efficiency, reducing runtime by 95\% on RAWFC (23 minutes 36 seconds compared to 454 minutes 12 seconds) and by 92\% on LIAR-RAW (134 minutes 14 seconds compared to 1692 minutes 23 seconds), showcasing its effectiveness across varying dataset sizes. On the RAWFC dataset, DeReC achieves an F1 score of 65.58\%, surpassing the state-of-the-art method L-Defense (61.20\%). Our results demonstrate that carefully engineered retrieval-based systems can match or exceed LLM performance in specialized tasks while being significantly more practical for real-world deployment.
\end{abstract}

\usetikzlibrary{positioning,shapes,arrows}


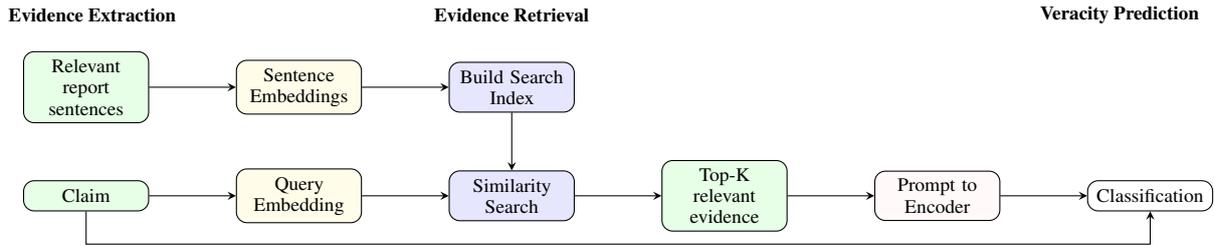
\begin{figure*}[t]
\centering
\begin{tikzpicture}[
    scale=0.8,
    transform shape,
    node distance = 0.7cm and 1.2cm,
    font=\rmfamily,
    box/.style = {
        draw,
        rounded corners,
        minimum width=1.4cm,
        minimum height=0.5cm,
        text width=1.8cm,
        align=center,
        font=\footnotesize
    },
    greenbox/.style = {box, fill=green!10},
    yellowbox/.style = {box, fill=yellow!10},
    bluebox/.style = {box, fill=blue!10},
    pinkbox/.style = {box, fill=pink!10},
    whitebox/.style = {box, fill=white},
    arrow/.style = {->, >=stealth, thin}
]
\node[font=\footnotesize\bfseries] at (0.1,3.0) {Evidence Extraction};
\node[font=\footnotesize\bfseries] at (7,3.0) {Evidence Retrieval};
\node[font=\footnotesize\bfseries] at (17,3.0) {Veracity Prediction};

\node[greenbox] (sentences) at (0,1.8) {Relevant report sentences};
\node[yellowbox] (embeddings) at (3.5,1.8) {Sentence Embeddings};
\node[bluebox] (search) at (7,1.8) {Build Search Index};

\node[greenbox] (claim) at (0,0) {Claim};
\node[yellowbox] (query) at (3.5,0) {Query Embedding};
\node[bluebox] (similarity) at (7,0) {Similarity Search};
\node[greenbox] (topk) at (10.5,0) {Top-K relevant evidence};
\node[pinkbox] (prompt) at (14,0) {Prompt to Encoder};
\node[whitebox] (output) at (17.5,0) {Classification};

\draw[arrow] (sentences) -- (embeddings);
\draw[arrow] (embeddings) -- (search);
\draw[arrow] (claim) -- (query);
\draw[arrow] (query) -- (similarity);
\draw[arrow] (search) -- (similarity);
\draw[arrow] (similarity) -- (topk);
\draw[arrow] (topk) -- (prompt);
\draw[arrow] (prompt) -- (output);

\draw[arrow] (claim) -- ++(0,-0.8) -| (output);

\end{tikzpicture}
\caption{DeReC: Three-Stage Pipeline for Evidence-Based Fact Verification.}
\label{fig:evidence-flow}
\end{figure*}

\section{Introduction}
The exponential growth of misinformation across digital platforms presents an urgent challenge to information integrity and societal discourse \cite{guo2022survey}. While recent advances in automated fact-verification systems have shown promise in addressing this challenge \cite{wang2024explainable,yue-etal-2024-retrieval,zhang-gao-2023-towards,yang2022coarse}, current approaches face significant limitations in both computational efficiency and verification reliability \cite{su2023adapting}. 

Recent work in automated fact-checking and in particular, state-of-the-art systems heavily rely on LLMs to generate natural language explanations that justify verification decisions \cite{wang2024explainable,zhang2023towards,yang2022coarse}. While these approaches have demonstrated impressive capabilities in reasoning about complex claims, they face three critical challenges: The computational demands of running inference with large models make real-time fact checking impractical \cite{tang2024minicheck}. LLM-generated explanations frequently contain hallucinations or factual inconsistencies that compromise verification reliability \cite{wang2024factuality}, and the generated rationales often lack direct grounding in verifiable evidence sources \cite{huang2023survey,su2023fake,yao2023llm,chen2024factchd}. Such limitations motivate the development of alternative strategies that prioritize both efficiency and transparency. 

In this work, we introduce DeReC (Dense Retrieval Classification), an evidence-enhanced hybrid framework that directly incorporates retrieved textual evidence into the fact-checking process. DeReC leverages sentence embeddings and Facebook AI Similarity Search (FAISS) \cite{douze2024faiss} to extract pertinent evidence from source documents, which is then integrated with the claim to form a robust input for downstream classification. By grounding predictions in actual evidence, we achieve both improved verification accuracy while significantly reducing computational overhead compared to LLM-based approaches. Unlike traditional Retrieval-Augmented Generation (RAG) systems that use retrieved content to enhance LLM prompts, DeReC directly grounds verification decisions in relevant evidence through efficient similarity search and targeted classification. Our experimental results demonstrate that this evidence-centric approach exceeds the performance of more complex LLM-based systems, while maintaining faster inference times and lower resource requirements. The code is available publicly. \footnote{Source code available at \url{https://github.com/alamgirqazi/DeReC}}

Our contribution can be summarized as follows: 
\begin{enumerate}
    \item We propose DeReC, a light-weight dense-retrieval-classification framework that combines advanced text embeddings with a specialized classifier to directly ground claims in factual evidence, achieving high verification accuracy without LLM-based rationale generation.
    
    \item We demonstrate that general-purpose text embeddings combined with dense retrieval can effectively replace LLM-based approaches in specialized tasks like fact verification, achieving better accuracy with significantly lower computational overhead (1.5B/137M parameters vs typical 7B+ LLM approaches).
      
    \item Empirical evaluations reveal state-of-the-art results on two datasets with an F1 score of 65.58\% on RAWFC and 33.13\% on LIAR-RAW.

\end{enumerate}

\begin{table*}[t]
  \centering
  \begin{tabular}{l|ccc|ccc}
    \hline
    \textbf{Metric} & \multicolumn{3}{c|}{\textbf{RAWFC}} & \multicolumn{3}{c}{\textbf{LIAR-RAW}} \\
    \cline{2-4} \cline{5-7}
    & Train & Val & Test & Train & Val & Test \\
    \hline
    Number of Claims & 1,612 & 200 & 200 & 10,065 & 1,274 & 1,251 \\
    Number of Reports & 33,862 & 4,127 & 4,278 & 114,721 & 18,243 & 21,408 \\
    Total Sentences & 248,343 & 31,191 & 31,453 & 626,573 & 102,147 & 118,449 \\
    Avg Sentences/Claim & 154.06 & 155.96 & 157.26 & 62.25 & 80.18 & 94.68 \\
    \hline
  \end{tabular}
  \caption{\label{tab:dataset-analysis}
    Analysis of dataset splits across LIAR-RAW and RAWFC datasets.
  }
\end{table*}

\section{Related Work}

\subsection{Fact Verification and Fake News Detection}

One of the earliest works on automatic fake news detection was introduced by \citet{vlachos2014fact}, who formally defined the fact-checking task, compiled a dataset from two popular fact-checking websites, and evaluated K-Nearest Neighbors classifiers for this purpose. \citet{popat2018declare} introduced an end-to-end neural network model for debunking fake news and false claims. It employs evidence and counter-evidences extracted from the web to support or refute a claim. 

The TI-CNN (Text and Image information based Convolutional Neural Network) model introduced in \citet{yang2018ti} leverages convolutional architecture to process entire inputs simultaneously, enabling faster training compared to sequential models like LSTMs and other RNNs. \citet{nasir2021fake} proposed a novel hybrid deep learning model that combines convolutional and recurrent neural networks for fake news classification.

\citet{shu2019defend} utilizes GRU-based model for veracity prediction with explanations. \citet{ma2019sentence} represents each sentence based on sentence-level coherence and semantic conflicts with the claim. \citet{kotonya2020explainable_a} uses SentenceBERT (SBERT) for encoding and detects fake news based on the top-K ranked sentences. \citet{atanasova2024generating} detects fake news independently or jointly with explanations in the multi-task set-up.

Current state-of-the-art systems frequently employ LLMs to generate natural language explanations for fact-checking decisions. \citet{yang2022coarse} proposed CofCED, a novel coarse-to-fine cascaded neural network for fake news detection that leverages the "wisdom of crowds" through raw media reports. \citet{shi2024generate} introduces a "generate-then-ground" framework for multi-hop question answering, where LLMs first generate answers to simplified sub-questions and then validate and correct these answers using retrieved external documents.

\subsection{Retrieval-Augmented Frameworks for Fact Verification}

Retrieval-Augmented Generation (RAG) has become an effective method for augmenting LLMs by integrating external retrieval mechanisms. Instead of relying solely on in-model knowledge, RAG enables models to retrieve relevant information from external documents during generation \cite{lewis2020retrieval}.  This approach has shown promise in many areas including open-domain question answering and dialogue systems \cite{izacard-grave-2021-leveraging}. Different types of RAG systems have been developed \cite{gao2023retrieval}, going from the original
naive RAG (simple structure of a retriever and a generator) \cite{lewis2020retrieval} to more advanced or modular RAG such as RA-DIT \cite{lin2023ra}.

In this paper, we adapts core principles from Retrieval-Augmented Generation (RAG) but replaces the generation component with efficient classification. Instead of augmenting an LLM's context for generation, we merge the extracted evidence with the input claim to create a robust, evidence-grounded input for a classifier. 

\subsection{Text Embeddings for Retrieval}

The evolution of text embedding models has fundamentally transformed information retrieval in natural language processing. Traditional approaches relied on sparse vector representations and lexical matching techniques like TF-IDF and BM25 \cite{robertson2009probabilistic}. 

Recent advancements in LLMs have significantly shifted the focus towards embedding models that rely primarily on decoder-only architectures \cite{liu2019roberta,li2024llama2vec}. These LLM-based embedding models have demonstrated remarkable improvements in in-domain accuracy and generalization, particularly when trained using supervised learning approaches \cite{wang2023improving}.

Recent advances in sentence embedding models have enabled more efficient and accurate retrieval for language tasks. While early approaches relied on sparse retrieval methods or basic transformer encoders, newer embedding models like \textit{Alibaba-NLP/gte-Qwen2-1.5B-instruct} have demonstrated superior performance in semantic search and retrieval tasks \cite{hui2024qwen2}. These models, trained on massive text pairs and optimized for similarity learning, provide dense vector representations that better capture semantic relationships between texts \cite{li2024making,nussbaum2024nomic}.

In this paper, we utilized two embedding models. The first is \textit{Alibaba-NLP/gte-Qwen2-1.5B-instruct} \cite{li2023towards}, a 1.5B parameter model that achieves strong performance through instruction tuning and contrastive learning. The second is \textit{nomic-ai/nomic-embed-text-v1.5} \cite{nussbaum2024nomic}, a more compact 137M parameter model that leverages Matryoshka representation learning to maintain high performance despite its reduced size. Both models demonstrate that effective dense retrieval can be achieved without the computational overhead of full-scale LLMs, making them particularly suitable for practical applications in fact verification.

\section{Method and Overall Architecture}

We present an integrated retrieval and classification architecture for automated fact verification that improves upon existing LLM-based methods. Our framework consists of three key components: evidence extraction using dense embeddings, evidence retrieval through FAISS-based similarity search, and veracity prediction using a specialized classifier.

\subsection{Evidence Extraction}

The evidence extraction phase involves processing the corpus of raw media reports to identify and represent potential evidence sentences. In this context, "extraction" refers to the process of transforming raw text from source documents into structured vector representations that can be efficiently retrieved and compared with claims.
Given a claim $c$ and a set of evidence sentences $\mathcal{E} = \{e_1, ..., e_n\}$, we employ dense embedding models to generate efficient vector representations. The embedding model can be formally defined as a function:
\begin{equation}
f: \mathcal{X} \rightarrow \mathbb{R}^d
\end{equation}
that maps any text sequence from the input space $\mathcal{X}$ to a d-dimensional real-valued vector space. For each input text $x$, the model generates a dense vector representation:
\begin{equation}
\mathbf{h}_x = f(x) \text{ where } \mathbf{h}_x \in \mathbb{R}^d
\end{equation}
where $d$ is the dimension of the embedding space. The embedding models are trained using contrastive learning objectives to ensure that semantically similar texts are mapped to nearby points in the embedding space. The similarity between two embeddings is computed using cosine similarity.

\subsection{Evidence Retrieval}

Using the dense vector representations generated during the evidence extraction stage (Section 3.1), We encode the original claim using the same embedding model and use FAISS for generating an inner product index optimized for cosine similarity search with normalized vectors. We configure FAISS to retrieve the top ten sentences most relevant to the claim. These sentences are then used in the veracity prediction module for final classification.
We utilize FAISS (Facebook AI Similarity Search) for efficient similarity search over the dense embeddings \citep{douze2024faiss}. FAISS is an efficient library for similarity search and clustering of dense vector space. FAISS constructs an optimized index structure $\mathcal{I}$ that supports fast nearest neighbor search over large collections of vectors.
Given the claim embedding $\mathbf{h}_c = f(c)$ and the set of evidence embeddings $\mathcal{H} = \{\mathbf{h}_1, ..., \mathbf{h}_n\}$ where $\mathbf{h}_i = f(e_i)$, we build a FAISS IndexFlatIP index optimized for inner product similarity search with normalized vectors:
\begin{itemize}
\item Vector normalization: $\bar{\mathbf{h}}_i = \frac{\mathbf{h}_i}{|\mathbf{h}_i|}$ for all vectors
\item Index construction: $\mathcal{I}.\text{add}(\bar{\mathcal{H}})$ where $\bar{\mathcal{H}}$ contains normalized vectors
\item Search: $\mathcal{I}.\text{search}(\bar{\mathbf{h}}_c, k)$ returns top-$k$ nearest neighbors
\end{itemize}
For normalized vectors, inner product corresponds to cosine similarity:
\begin{equation}
\bar{\mathbf{h}}_c^\top\bar{\mathbf{h}}_i = \cos(\mathbf{h}_c, \mathbf{h}_i)
\end{equation}
The index supports sub-linear $\mathcal{O}(\log n)$ search complexity compared to linear $\mathcal{O}(n)$ for exhaustive search.

For each claim, we retrieve the top-$k$ most relevant evidence sentences using cosine similarity. Based on empirical validation on the development set, we set $k=10$ as it provides an optimal balance between computational efficiency and evidence coverage.

\subsection{Evidence-Enhanced Veracity Prediction}

The Veracity Prediction component employs DeBERTa-v3-large \cite{he2020deberta} fine-tuned for multi-class veracity prediction. Given a claim $c$ and retrieved evidence $\mathcal{E}$, we construct the input sequence:
\begin{equation}
x = [\text{CLS}] ; c ; [\text{SEP}] ; e_1 ; [\text{SEP}] ; ... ; [\text{SEP}] ; e_k ; [\text{SEP}]
\end{equation}
where $k$ is the number of retrieved evidence pieces.
The model computes contextual representations:
\begin{equation}
\mathbf{H} = \text{DeBERTa}(x) \in \mathbb{R}^{d \times L}
\end{equation}
where $L$ is the sequence length and $d$ is the hidden dimension.
The [CLS] token representation is used for classification:
\begin{equation}
\mathbf{h}_{[\text{CLS}]} = \mathbf{H}_0 \in \mathbb{R}^d
\end{equation}
\begin{equation}
\mathbf{y} = \text{softmax}(\mathbf{W}\mathbf{h}_{[\text{CLS}]} + \mathbf{b})
\end{equation}
where $\mathbf{W} \in \mathbb{R}^{c \times d}$ and $\mathbf{b} \in \mathbb{R}^c$ are learned parameters, and $c$ is the number of classes.
During training, we minimize the cross-entropy loss:
\begin{equation}
\mathcal{L} = -\sum_i y_i\log(\hat{y}_i)
\end{equation}
where $y$ is the ground truth label and $\hat{y}$ is the predicted probability distribution.

The model was fine-tuned on the training splits of the LIAR-RAW and RAWFC datasets (described in Section 4.2), with separate models trained for each dataset to account for their different label distributions.

The classification component implements a encoder-based transformer architecture DeBERTa-v3-large \cite{he2020deberta} optimized for multi-class veracity prediction. DeBERTa-v3-large improves the BERT and RoBERTa models using two novel techniques. The first is the disentangled attention mechanism, where each word is represented using two vectors that encode its content and position. Second, an enhanced mask decoder is used to incorporate absolute positions in the decoding layer to predict the masked tokens in model pre-training.

\subsection{Computational Efficiency}

DeReC achieves significant computational advantages through its three-stage architecture. For a sequence of length $l$ and corpus size $s$, the computational complexity can be broken down by stage:

\begin{itemize}
    \item
The embedding stage utilizes a parameter-efficient model (1.5B / 137M parameters) with linear complexity $O(l)$ for processing input text.
\item
The FAISS-based similarity search stage achieves logarithmic complexity $O(\log s)$, enabling efficient retrieval even for large document collections.
\item
The classification stage requires only a single forward pass through a encoder transformer model with complexity $O(l)$.

\end{itemize}

This results in a total computational complexity of $O(l + \log s)$, which compares favorably to LLM-based approaches requiring $O(n \times l^2)$ operations for a model with $n$ layers. Memory requirements are similarly reduced from $M_{llm} = O(p_{llm} \times b + l^2)$ for LLM approaches (where $p_{llm}$ is typically 7B+ parameters) to $M_{derec} = O(p_{emb} \times b + s)$ for our method (where $p_{emb}$ is 1.5B / 137M parameters and $b$ is bits per parameter).

These theoretical improvements yield substantive practical benefits: the elimination of computationally intensive text generation operations, a reduced memory complexity that scales linearly with corpus size rather than quadratically with sequence length, and the capacity for deployment on commodity hardware while maintaining competitive performance metrics.

 \section{Experiments and Results}
 
Our framework achieves state-of-the-art results on both LIAR-RAW and RAWFC benchmarks, outperforming all baseline models in terms of F1 score, precision and recall. 

\subsection{Experimental Setup}

We evaluate our framework on two extensive benchmarks: LIAR-RAW and RAWFC. Our experimental setup includes training the dense retriever and veracity prediction models separately, followed by end-to-end fine-tuning to optimize performance. All experiments are conducted on a single NVIDIA A40 GPU with PyTorch framework.

We employ two different embedding models for generating efficient dense embeddings from the sentences. \textit{Alibaba-NLP/gte-Qwen2-1.5B-instruct} is a 1.5B embedding model that provides efficient embeddings for sentences. It has shown strong performance on the MTEB (Massive Text Embedding Benchmark, \citealp{muennighoff2022mteb}). The second embedding model we used is a much smaller 137M model \textit{nomic-ai/nomic-embed-text-v1.5}.

 The models were selected based on comprehensive evaluation across the MTEB suite, offering an optimal balance between embedding quality and computational efficiency.
 
For the retriever component, we employ \textit{Alibaba-NLP/gte-Qwen2-1.5B-instruct} as our primary embedding model. Document retrieval utilizes FAISS with an inner product index optimized for cosine similarity search with normalized vectors. The embeddings are generated through our model and added to the FAISS index for efficient similarity search. For classification, we utilize DeBERTa-v3-large with a maximum sequence length of 512 tokens.

\subsection{Datasets}

\begin{table}[t]
  \centering
  \begin{tabular}{l|cc}
    \hline
    \textbf{Veracity Label} & \textbf{RAWFC} & \textbf{LIAR-RAW} \\
    \hline
    pants-fire & - & 1,013 \\
    false & 646 & 2,466 \\
    barely-true & - & 2,057 \\
    half-true & 671 & 2,594 \\
    mostly-true & - & 2,439 \\
    true & 695 & 2,021 \\
    \hline
    \multicolumn{1}{l|}{\textbf{Total Claims}} & 2,012 & 12,590 \\
    \multicolumn{1}{l|}{\textbf{Veracity Labels}} & 3 & 6 \\
    \hline
  \end{tabular}
  \caption{\label{tab:dataset-stats}
    Distribution of veracity labels across RAWFC and LIAR-RAW datasets. 
  }
\end{table}

We conducted our evaluation using two extensively documented datasets: RAWFC and LIAR-RAW \cite{yang2022coarse}, with their detailed characteristics and distributions presented in Table 1 and their veracity labels detailed on Table 2.
LIAR-RAW is an expanded version of the LIAR-PLUS dataset \cite{alhindi2018your}. The dataset employs a fine-grained six-class classification scheme: pants-fire, false, barely-true, half-true, mostly-true, and true. Each claim in the dataset is accompanied by relevant raw news reports and documents that were collected during the dataset's creation.

The RAWFC dataset \cite{yang2022coarse}, derived from Snopes.com claims, implements a more condensed three-class classification system (false, half, true). The dataset includes claims along with their associated raw reports retrieved using claim keywords. 
\begin{table*}[t]
  \centering
  \begin{tabular}{l|ccc|ccc}
    \hline
    & \multicolumn{3}{c|}{RAWFC} & \multicolumn{3}{c}{LIAR-RAW} \\
    \cline{2-7}
    & P & R & F1 & P & R & F1 \\
    \hline \hline
    \multicolumn{7}{l}{\textit{Traditional approach}} \\
    \hline
    dEFEND \cite{shu2019defend} & 44.90 & 43.20 & 44.00 & 23.00 & 18.50 & 20.50 \\
    SentHAN \cite{ma2019sentence} & 45.70 & 45.50 & 45.60 & 22.60 & 20.00 & 21.20 \\
    SBERT-FC \cite{kotonya2020explainable_b,kotonya2020explainable_a} & 51.10 & 46.00 & 48.40 & 24.10 & 22.10 & 23.10 \\
    CofCED \cite{yang2022coarse} & 53.00 & 51.00 & 52.00 & 29.50 & 29.60 & 29.50 \\
    GenFE \cite{atanasova2024generating} & 44.29 & 44.74 & 44.43 & 28.01 & 26.16 & 26.49 \\
    GenFE-MT \cite{atanasova2024generating} & 45.64 & 45.27 & 45.08 & 18.55 & 19.90 & 15.15 \\
    \hline
    \hline
    \multicolumn{7}{l}{\textit{LLM-based approach}} \\
    \hline
    
    FactLLaMA \cite{cheung2023factllama} & 53.76 & 54.00 & 53.76 & 29.98 & 31.57 & 32.32 \\
    FactLLaMA$_{\text{know}}$ \cite{cheung2023factllama} & 55.65 & 55.50 & 56.11 & 30.44 & 32.05 & 32.46 \\
    L-Defense$_{\text{ChatGPT}}$ \cite{wang2024explainable} & 61.72 & 61.91 & 61.20 & 30.55 & \underline{32.20} & 30.53 \\
    L-Defense$_{\text{LLaMA2}}$ \cite{wang2024explainable} & 60.95 & 60.00 & 60.12 & 31.63 & 31.71 & 31.40 \\
    \hline
    \hline
    \multicolumn{7}{l}{\textit{Ours}} \\
    \hline
    DeReC-qwen & \textbf{65.58} & \underline{64.56} & \underline{64.60} & \textbf{35.94} & \textbf{32.24} & \textbf{33.13} \\
    DeReC-nomic & \underline{64.48} & \textbf{65.57} & \textbf{64.61} & \underline{33.19} & 31.50 & \underline{31.79} \\
    \hline
  \end{tabular}
  \caption{Performance comparison across RAWFC and LIAR-RAW datasets. Best scores are in \textbf{bold} and second-best scores are \underline{underlined} for each metric.}
  \label{tab:combined-results}
\end{table*}

\subsection{Baseline Models}

We compare our approach against state-of-the-art traditional and LLM based approaches including L-Defense \cite{wang2024explainable} without external sources. L-Defense employs a three-stage framework: 1) an evidence extraction module that uses RoBERTa-base to split and rank evidence into competing true and false narratives, 2) a prompt-based reasoning module utilizing LLMs (either ChatGPT or LLaMA2-7B) to generate explanations for both perspectives, and 3) a defense-based inference module with RoBERTa-large that determines the final veracity prediction.
For fair comparison, all baselines were evaluated in their supervised settings, using the same training data as our approach. Models like L-Defense and FactLLaMA, while capable of zero-shot inference, were fine-tuned on the task-specific data to ensure comparable evaluation conditions.

\begin{table*}[t] 
\centering
\vspace{0.2cm}
\resizebox{\textwidth}{!}{%
\begin{tabular}{l|l|c|c|c}
\hline
\textbf{Dataset} & \textbf{Step} & DeReC-nomic & DeReC-qwen & L-Defense$_{\text{LLaMA2}}$ \\
\hline
RAWFC
& Evidence Extraction    & 3m 50s  & 35m 15s & 61m 39s \\
& Evidence Retrieval    & 2m 2s  & 7m 26s  & - \\
& LLM-generated Explanations   & -  & - & 381m 31s \\
& Veracity Prediction   & 17m 44s & 21m 30s & 11m 2s \\
\cline{2-5}
& \textbf{Total Runtime} & \textbf{23m 36s} & \textbf{64m 11s} & \textbf{454m 12s} \\
\hline
LIAR-RAW
& Evidence Extraction   & 9m 17s  & 89m 21s & 185m 59s \\
& Evidence Retrieval    & 30m 12s & 45m 13s & - \\
& LLM-generated Explanations   & -  & - & 1466m 8s \\
& Veracity Prediction   & 94m 45s & 89m 53s & 40m 16s \\
\cline{2-5}
& \textbf{Total Runtime} & \textbf{134m 14s} & \textbf{254m 48s} & \textbf{1692m 23s} \\
\hline
\end{tabular}%
}
\vspace{0.3cm}  %
\caption{Step-wise runtime breakdown (in minutes and seconds) for different models.}
\label{tab:runtime}
\end{table*}

\begin{figure*}[t]
    \centering
    \begin{tikzpicture}
    \begin{axis}[
        width=0.85\columnwidth,
        height=4cm,
        ybar=4pt,
        bar width=9pt,
        symbolic x coords={DeReC-nomic, DeReC-qwen, L-Defense},
        xtick=data,
        xticklabels={DeReC-nomic, DeReC-qwen, L-Defense\textsubscript{LLaMA2}},
        ymin=0, ymax=2000,
        ylabel={Runtime (min)},
        ylabel style={font=\small},
        enlarge x limits=0.15,
        xticklabel style={font=\small, text height=1.5ex, text depth=0.5ex},
        yticklabel style={font=\small},
        ymajorgrids=true,
        grid style={dotted, opacity=0.3},
        major grid style={white!90!black},
        axis lines=box,
        legend style={
            at={(0.5,-0.3)},
            anchor=north,
            legend columns=2,
            font=\small,
            /tikz/every even column/.append style={column sep=0.2cm}
        },
        legend entries={RAWFC, LIAR-RAW},
        tick align=outside,
        scale only axis,
        clip=false,
    ]
        \addplot[fill={rgb:red,0.1;green,0.4;blue,0.7}, opacity=0.85] coordinates {
            (DeReC-nomic, 23.60)
            (DeReC-qwen, 64.18)
            (L-Defense, 454.21)
        };
        \node[font=\tiny, yshift=4pt, xshift=-6pt] at (axis cs:DeReC-nomic, 23.60) {23.60};
        \node[font=\tiny, yshift=4pt, xshift=-6pt] at (axis cs:DeReC-qwen, 64.18) {64.18};
        \node[font=\tiny, yshift=4pt, xshift=-6pt] at (axis cs:L-Defense, 454.21) {454.21};
        
        \addplot[fill={rgb:red,0.85;green,0.37;blue,0.01}, opacity=0.85] coordinates {
            (DeReC-nomic, 134.23)
            (DeReC-qwen, 254.80)
            (L-Defense, 1692.38)
        };
        \node[font=\tiny, yshift=4pt, xshift=6pt] at (axis cs:DeReC-nomic, 134.23) {134.23};
        \node[font=\tiny, yshift=4pt, xshift=6pt] at (axis cs:DeReC-qwen, 254.80) {254.80};
        \node[font=\tiny, yshift=4pt, xshift=6pt] at (axis cs:L-Defense, 1692.38) {1692.38};
    \end{axis}
    \end{tikzpicture}
    \caption{Complete pipeline runtime comparison (in minutes) on RAWFC and LIAR-RAW datasets.}
    \label{fig:total_runtime}
\end{figure*}
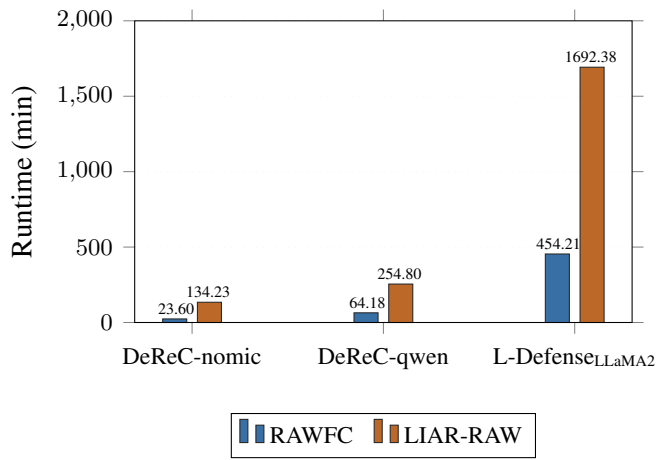

\subsection{Results and Analysis}

We evaluate our framework using two variants: DeReC-qwen, which employs the 1.5B parameter \textit{Alibaba-NLP/gte-Qwen2-1.5B-instruct} embedding model, and DeReC-nomic, which utilizes the 137M \textit{nomic-ai/nomic-embed-text-v1.5} model. Both variants demonstrate strong performance across datasets, with DeReC-qwen achieving state-of-the-art results which DeReC-nomic getting better results compared to all previous approaches for both datasets except for Recall in L-Defense (ChatGPT) for LIAR-RAW dataset.
On the RAWFC dataset, our models achieve strong F1 scores, with DeReC-nomic reaching 64.61\% and DeReC-qwen achieving 64.60\%. Both significantly outperform previous leading methods, including  L-Defense$_{\text{ChatGPT}}$ (\textbf{61.20\%}) and  L-Defense$_{\text{LLaMA2}}$ (\textbf{60.12\%}). The performance improvement is particularly significant given that our method requires substantially less computational resources by eliminating LLM-based explanation generation. The model demonstrates robust performance across all metrics, with precision reaching \textbf{65.58\%} and recall achieving \textbf{64.56\%}, indicating balanced and consistent prediction capabilities.
For the LIAR-RAW dataset, which presents a more challenging six-class classification task, our method achieves an F1 score of \textbf{33.13\%}, surpassing both variants of L-Defense and traditional approaches. The improvement is particularly pronounced in precision (\textbf{35.94\%}), suggesting that our evidence retrieval mechanism effectively reduces false positives.

For the more challenging LIAR-RAW dataset, which requires six-class classification, DeReC-qwen attains an F1 score of \textbf{33.13\%}, outperforming both variants of L-Defense and traditional baseline approaches. The notable improvement in precision (\textbf{35.94\%}) suggests that our evidence retrieval mechanism effectively minimizes false positives, leading to more reliable classification outcomes.

Our lightweight variant, DeReC-nomic, demonstrates comparable effectiveness on the RAWFC dataset, achieving an \textit{F\textsubscript{1}} score of \textbf{64.61\%}. However, it shows moderate performance degradation on the more complex LIAR-RAW dataset, suggesting that the additional capacity of the DeReC-qwen model may be beneficial for more nuanced classification tasks.

\subsection{Runtime Analysis}
As shown in Table 4 and Figure 2, our framework achieves substantial runtime improvements compared to explainable-generating LLM-based approaches. All runtime experiments were conducted using a single NVIDIA A40 GPU. On the RAWFC dataset, DeReC-nomic completes the entire pipeline in 23 minutes and 36 seconds, representing a 95\% reduction in total runtime compared to L-Defense$_{\text{LLaMA2}}$ (454 minutes and 12 seconds). The larger DeReC-qwen variant maintains significant efficiency advantages while offering enhanced performance, completing processing in 64 minutes and 11 seconds.
The step-wise runtime breakdown reveals that the most substantial efficiency gains come from eliminating LLM-based explanation generation, which consumes 381 minutes and 31 seconds (84\%) of L-Defense's total runtime on RAWFC. Our evidence extraction and retrieval pipeline, in contrast, requires only 5 minutes and 52 seconds for DeReC-nomic and 42 minutes and 41 seconds for DeReC-qwen. This dramatic reduction is achieved while maintaining superior classification performance, demonstrating that expensive generative inference is not necessary for effective fact verification.
The efficiency advantages scale consistently to larger datasets. On LIAR-RAW, which contains approximately 6 times more claims than RAWFC, DeReC-nomic completes processing in 134 minutes and 14 seconds compared to L-Defense's 1692 minutes and 23 seconds. The primary bottleneck in the L-Defense approach is the LLM explanation generation step, requiring 1466 minutes and 8 seconds (87\% of total runtime). Our retrieval-based architecture eliminates this bottleneck entirely, with combined evidence extraction and retrieval taking only 39 minutes and 29 seconds for DeReC-nomic and 134 minutes and 34 seconds for DeReC-qwen.
These runtime improvements have significant practical implications for real-world deployment. While LLM-based approaches require substantial GPU resources for batch processing, our framework's efficiency enables near real-time fact verification on consumer hardware. The modular nature of our architecture also allows for straightforward scaling through parallel processing of the evidence extraction and retrieval stages, offering a clear path to handling larger evidence corpora.

 \section{Discussion}

Our experimental results reveal several key insights about the relationship between evidence retrieval and fact verification. The performance improvements achieved by our hybrid architecture suggest important implications for future development of automated fact-checking systems.

The memory footprint differential between these approaches is substantial. LLM-based methods must maintain the full model parameters in GPU memory while also allocating space for attention computations that scale quadratically with sequence length. Additionally, these models require KV-cache memory for generation \cite{chowdhery2023palm}. 

Traditional approaches utilizing LLMs such as ChatGPT (175B+ parameters) or LLaMA2 (7B parameters) for explanation generation face significant computational challenges. The fundamental bottleneck lies in the autoregressive nature of text generation, which necessitates sequential processing with quadratic complexity $O(n^2)$ for generating n tokens. These models require substantial GPU memory allocation due to their massive parameter counts. Moreover, methods like L-Defense require multiple LLM calls per claim to generate competing explanations, further amplifying the computational overhead.

Our retrieval-based approach fundamentally refactors this paradigm by eliminating the need for explanation generation entirely. The architecture employs a significantly smaller embedding model (\textit{nomic-ai/nomic-embed-text-v1.5}, 137M) requires only 0.5GB in FP32 precision which still beats most benchmarks while a slightly bigger model (\textit{Alibaba-NLP/gte-Qwen2-1.5B-instruct}, 1.5B parameters) requires only 6GB in FP32 precision. This model performs single-pass encoding with linear complexity $O(n)$, followed by efficient FAISS-based similarity search with sub-linear complexity $O(\log k)$ for k evidence sentences. The final classification step utilizes a lightweight DeBERTa-v3-large classifier (304M parameters) that requires only a single forward pass.

\section{Conclusion}
We present a hybrid retrieval-classification framework for fact verification that achieves state-of-the-art performance on the LIAR-RAW and RAWFC benchmarks. Our approach demonstrates that carefully engineered dense retrieval systems can match or exceed the performance of LLMs while significantly reducing computational overhead. The empirical results show that DeReC achieves a 95\% reduction in runtime while improving accuracy, challenging the assumption that LLM-based generation is necessary for effective fact verification.

Our findings have several important implications for the field of automated fact-checking. First, they demonstrate that efficient dense embeddings combined with targeted classification can effectively replace more complex LLM-based approaches in specialized tasks. Second, the dramatic reduction in computational requirements (from 7B+ parameters to 137M-1.5B) makes real-time fact verification more practically feasible for deployment in resource-constrained environments. Third, our results suggest that explicit rationale generation, while interpretable, may not be necessary for achieving high verification accuracy.

The modular nature of our architecture enables straightforward incorporation of improved embedding models as they become available. Our results suggest several promising research directions: investigating methods for dynamic evidence corpus updates, exploring techniques for handling multilingual verification scenarios, and developing lightweight explanation generation methods that maintain both computational efficiency and interpretability.

These findings contribute to the broader discussion about the role of large language models in practical applications, suggesting that targeted, efficient approaches may often be preferable to more computationally intensive general-purpose models. As misinformation continues to pose significant challenges to online discourse, frameworks like DeReC demonstrate how we can build more scalable and efficient solutions for automated fact verification.

 \section{Limitations}

 While our framework demonstrates strong performance, it is not without limitations. The quality of retrieval is heavily dependent on the evidence corpus; incomplete or biased corpora can lead to suboptimal results.
 Although our approach is more efficient than LLM-based methods, the FAISS index still requires significant memory for large-scale deployments. The index size scales linearly with the number of evidence sentences, which can create memory constraints for very large evidence corpora. While our approach prioritizes efficiency and recall, it does not generate natural language explanations for its decisions. This limitation may reduce its utility in contexts where detailed explanations are necessary for human review.
 
\section*{Acknowledgments}
This research is supported by the European Union’s Horizon Europe programme under grant agreement No 101135757, project AI4Debunk\footnote{\url{https://ai4debunk.eu}}. John P. McCrae is supported by Research Ireland under Grant Number SFI/12/RC/2289\_P2 Insight\_2, Insight SFI Centre for Data Analytics and Grant Number 13/RC/2106\_P2, ADAPT SFI Research Centre.

\bibliography{custom}

\end{document}